\documentclass[letterpaper, 10pt, conference]{ieeeconf}

\IEEEoverridecommandlockouts
\usepackage{graphicx}
\usepackage{amsmath}
\usepackage{amssymb}
\usepackage{mathnotation}
\usepackage{cite}
\usepackage{empheq}
\usepackage[final, defaultcolor=red]{changes}

\begin{document}

\title{Towards Geometric Motion Planning for High-Dimensional Systems: Gait-Based Coordinate Optimization and Local Metrics}

\author{Yanhao Yang, Capprin Bass, Ross L. Hatton %
\thanks{This work was supported in part by NSF Grants No. 1653220, 1826446, and 1935324.} %
\thanks{Y. Yang and R. L. Hatton are with the  Collaborative Robotics and Intelligent Systems (CoRIS) Institute at Oregon State University, Corvallis, OR USA. {\tt\small \{yangyanh, Ross.Hatton\}@oregonstate.edu}} %
\thanks{\added{C. Bass is with the AI Institute, 145 Broadway, Cambridge, MA USA. The bulk of his contributions to this work occurred while at the CoRIS Institute. {\tt\small cbass@theaiinstitute.com}}} %
}

\maketitle

\begin{abstract}

Geometric motion planning offers effective and interpretable gait analysis and optimization tools for locomoting systems. However, due to the curse of dimensionality in coordinate optimization, a key component of geometric motion planning, it is almost infeasible to apply current geometric motion planning to high-dimensional systems. In this paper, we propose a gait-based coordinate optimization method that overcomes the curse of dimensionality. We also identify a unified geometric representation of locomotion by generalizing various nonholonomic constraints into local metrics. By combining these two approaches, we take a step towards geometric motion planning for high-dimensional systems. We test our method in two classes of high-dimensional systems - low Reynolds number swimmers and free-falling Cassie - with up to 11-dimensional shape variables. The resulting optimal gait in the high-dimensional system shows better efficiency compared to that of the reduced-order model. Furthermore, we provide a geometric optimality interpretation of the optimal gait.

\end{abstract}

\section{Introduction}

Locomotion is one of the most fundamental actions of robots and animals. 
While animals can easily achieve locomotion by coordinating joints and muscles \cite{alexander2006principles}, reproducing locomotion in robots remains challenging, which is mainly due to the complex and multidimensional dynamics involved and interactions with various environments.

Geometric motion planning offers an effective and interpretable approach to locomotion analysis and gait design for various locomoting systems and environments. This approach utilizes the constraint curvature of the system to generate the ``corrected body velocity integral'' (cBVI) to approximate gait-induced displacement \cite{hatton2015nonconservativity}. Recent studies have combined cBVI with various power consumption metrics for gait optimization \cite{ramasamy2019geometry, hatton2022geometry}, and have generated practical gaits for robots \cite{sparks2022amoebainspired, rozaidi2023hissbot}. The accuracy of cBVI highly depends on the choice of coordinates \cite{bass2022characterizing}. Therefore, a fundamental step in geometric motion planning is coordinate optimization, aiming to minimize the approximation error \cite{hatton2011geometric}.

However, the coordinate optimization suffers from the curse of dimensionality. This challenge arises from the requirement to solve large-scale partial differential equations (PDEs) across the entire system's degrees of freedom (DOFs), which generally requires the use of the finite element method (FEM) \cite{hatton2011geometric}. Unfortunately, the complexity of the FEM grows exponentially with the number of dimensions. For this reason, applying geometric motion planning to high-dimensional systems becomes nearly impractical.

Although attempts have been made to address this issue via random sampling of the configuration space in a mesh-free manner, the number of samples required still increases exponentially with the dimensionality in order to achieve sufficiently accurate results \cite{chen2023integrable}. Other possible approaches are to find an approximate analytical solution for the optimal body coordinates by averaging the coordinates of the system's components \cite{hatton2009approximating} or aligning them with the principal axes of the system's composite inertia \cite{rollinson2012virtual, du2021meaningful}. \added{However, these approximations are only valid near the nominal configuration and are limited to inertia-dominated systems. Finally, although it is still possible to perform geometric motion planning without coordinate optimization \cite{melli2006motion}, it is only valid when the system has very little non-commutativity \cite{hatton2015nonconservativity, bass2022characterizing}.}

\begin{figure}[!t]
\centering
\includegraphics[width=\linewidth]{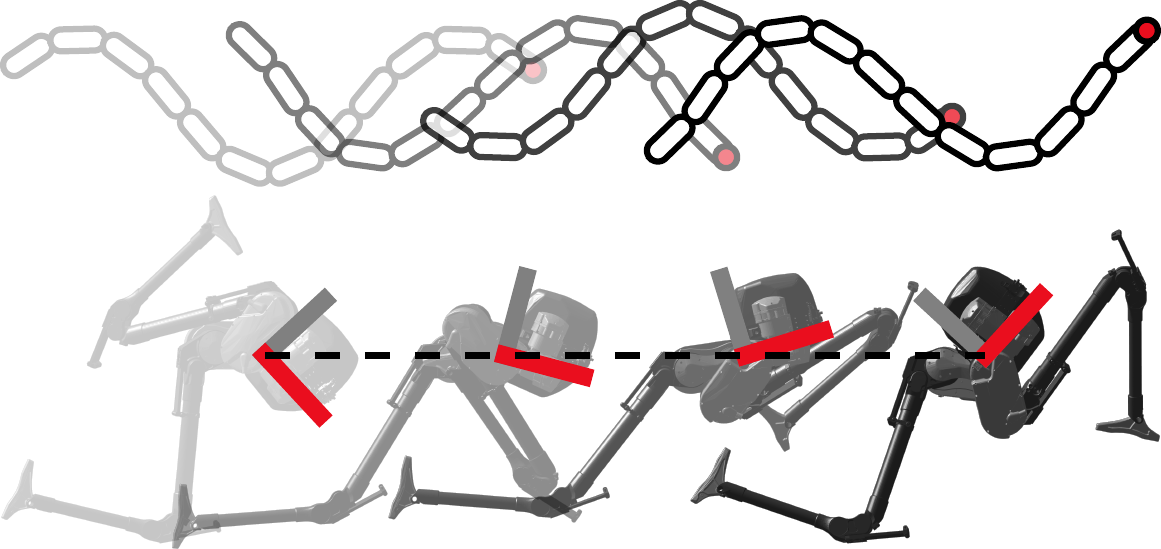}
\caption{Optimal gaits for the forward motion of a 12-link low Reynolds number swimmer and the pitch-rotational motion of a full model Cassie solved by the proposed high-dimensional geometric motion planning.}
\label{fig:intro}
\end{figure}

In this paper, we take a step towards geometric motion planning for high-dimensional systems by proposing a new method for coordinate optimization and a unified modeling framework. Specifically, the contribution contains:
\begin{itemize}
\item A gait-based coordinate optimization method, whose complexity scales linearly with the system dimension;
\item A unified geometric representation of locomotion summarizing the previous modeling of locomoting systems dominated by various nonholonomic constraints; and
\item Implementation of geometric motion planning on two different high-dimensional systems and their corresponding geometric optimality interpretation.
\end{itemize}

Specifically, the proposed approach was tested on two different types of high-dimensional systems: 1) drag-dominated low Reynolds number swimmer with 3, 6, 9, and 12 links of the same total length, and 2) inertia-dominated free-falling Cassie, a bipedal robot, with a 4-DOF reduced-order model 
and a 10-DOF full model. Fig.~\ref{fig:intro} shows that our method effectively finds optimal gaits for these systems that exploit the additional DOFs to increase efficiency\deleted{ with the same level of actuation effort}.

\section{Geometric Representation of Locomotion}

In this section, we propose a unified geometric representation of locomotion by generalizing various nonholonomic constraints into local metrics. We also review the geometric mechanics techniques that will be used in this paper, including local connections, constraint curvature, cBVI, and geometric gait optimization, all of which can be derived from the proposed unified geometric representation of locomotion.

The configuration space $\bundlespace$ of a multi-body locomoting system (the space of its generalized coordinates $\bundle$) can be divided into \emph{position} space $\fiberspace$, which locates the system in the world, and \emph{shape} space $\basespace$, which gives the relative arrangement of the system components.\footnote{Formally, this decoupling is embodied in a differential geometric structure called a \emph{fiber bundle}, where the \emph{shape} and \emph{position} of the system are respectively referred to as elements of the \emph{base} and \emph{fiber} spaces \cite{bloch2015nonholonomic}.} For example, the position of a 3-link low Reynolds number swimmer, as shown at the top of Fig.~\ref{fig:swimmer_gait}, can be its center of mass and central link orientation, $\fiber = (x,y,\theta) \in SE(2)$, and its shape is parameterized by the joint angles, $\base = (\alpha_{1},\alpha_{2})$.

\subsection{Local Metric Representation of Nonholonomic Constraints}

For a nonholonomic locomoting system, we can express its physics locally on the $\fstidx$-th component as a linear map $\pfaffian_{\fstidx}$ from the \added{body} velocity $\fibercirc_{\fstidx}$ in the constraint frame $\fiber_{\fstidx}$ to a relevant physical quantity, such as momentum or drag force. Formally, this linear map is a local metric of the motion of the system. 
Local metrics for common nonholonomic locomoting systems and environments are \cite{hatton2011geometric, hatton2013geometric}
\begin{subequations}
\begin{empheq}[left={\pfaffian_{\fstidx}=\empheqlbrace}]{align}
& \begin{bmatrix}\mass & \mass & I\end{bmatrix} \span (\text{Momentum}) & \label{eq:momentum} \\
& \begin{bmatrix}L & \dragcoeff L & \dragcoeff L^{3}/12\end{bmatrix} \span (\text{Drag force}) & \label{eq:drag} \\
& \begin{bmatrix}\mass & \mass & I\end{bmatrix} & (\text{Fluid-added momentum}) & \\
& \quad + \begin{bmatrix}\rho\pi b^{2} & \rho\pi a^{2} \span \rho(a^2-b^2)^2\end{bmatrix} \span \span \nonumber \\
& \begin{bmatrix}0 & 1 & 0\end{bmatrix} \span (\text{Nonslip}) & \\
& \begin{bmatrix}1 & 1 & 0\end{bmatrix}, \span (\text{Nonslip and nonslide}) &
\end{empheq}
\end{subequations}
\added{where the physical terms are defined as follows}: $\mass$ for mass, $I$ for moment of inertia, $\dragcoeff$ for the ratio between longitudinal and lateral drag coefficients, $L$ for the length of the elongated body, $\rho$ for fluid density, and $a$ and $b$ for the semimajor and semiminor axis lengths of the ellipse-shaped link. 

Using the system kinematics, we can transform these local metrics into an aggregate physical expression on the system that encodes the solution to the system's equations of motion\added{ and can represent physical rules like the conservation of momentum or force balance of the system}
\begin{eqalign}
\transpose{\jac_{\fiber}}\diag\left(\begin{bmatrix}\pfaffian_{1} & \cdots & \pfaffian_{\fstqty}\end{bmatrix}\right)\transpose{\begin{bmatrix}\transpose{\fibercirc}_{1} & \cdots & \transpose{\fibercirc}_{\fstqty}\end{bmatrix}} &= 0 \\ \transpose{\jac_{\fiber}}\pfaffian\begin{bmatrix}\jac_{\fiber} & \jac_{\base}\end{bmatrix}\transpose{\begin{bmatrix}\transpose{\fibercirc} & \transpose{\basedot}\end{bmatrix}} &= 0, \label{eq:nhc}
\end{eqalign}
where $\jac_{\fiber}$ and $\jac_{\base}$ are the Jacobians of the body velocity at each constraint frame with respect to the system's position velocity $\fibercirc$ and shape velocity $\basedot$, respectively,\footnote{$\inv{\Adj}_{\fiber_{\fstidx}}$ is an adjoint operator on the position space that combines cross-product and rotation matrix to transfer velocities and forces/moments between parts of the system. $\inv{\fiber}_{\fstidx}\grad \fiber_{\fstidx}$ represents the Jacobian matrix that relates the body velocity at each constraint frame to the shape velocity.}
\begin{eqalign}
\jac_{\fiber} &= \transpose{\begin{bmatrix}\transpose{\left(\inv{\Adj}_{\fiber_{1}}\right)} & \cdots & \transpose{\left(\inv{\Adj}_{\fiber_{\fstqty}}\right)}\end{bmatrix}} \\
\jac_{\base} &= \transpose{\begin{bmatrix}\transpose{\left(\inv{\fiber}_{1}\grad \fiber_{1}\right)} & \cdots & \transpose{\left(\inv{\fiber}_{\fstqty}\grad \fiber_{\fstqty}\right)}\end{bmatrix}}.
\end{eqalign}
\deleted{This aggregate physical expression can represent physical rules like the conservation of momentum or force balance of the system at the body frame.}

\begin{figure}[!t]
\centering
\includegraphics[width=\linewidth]{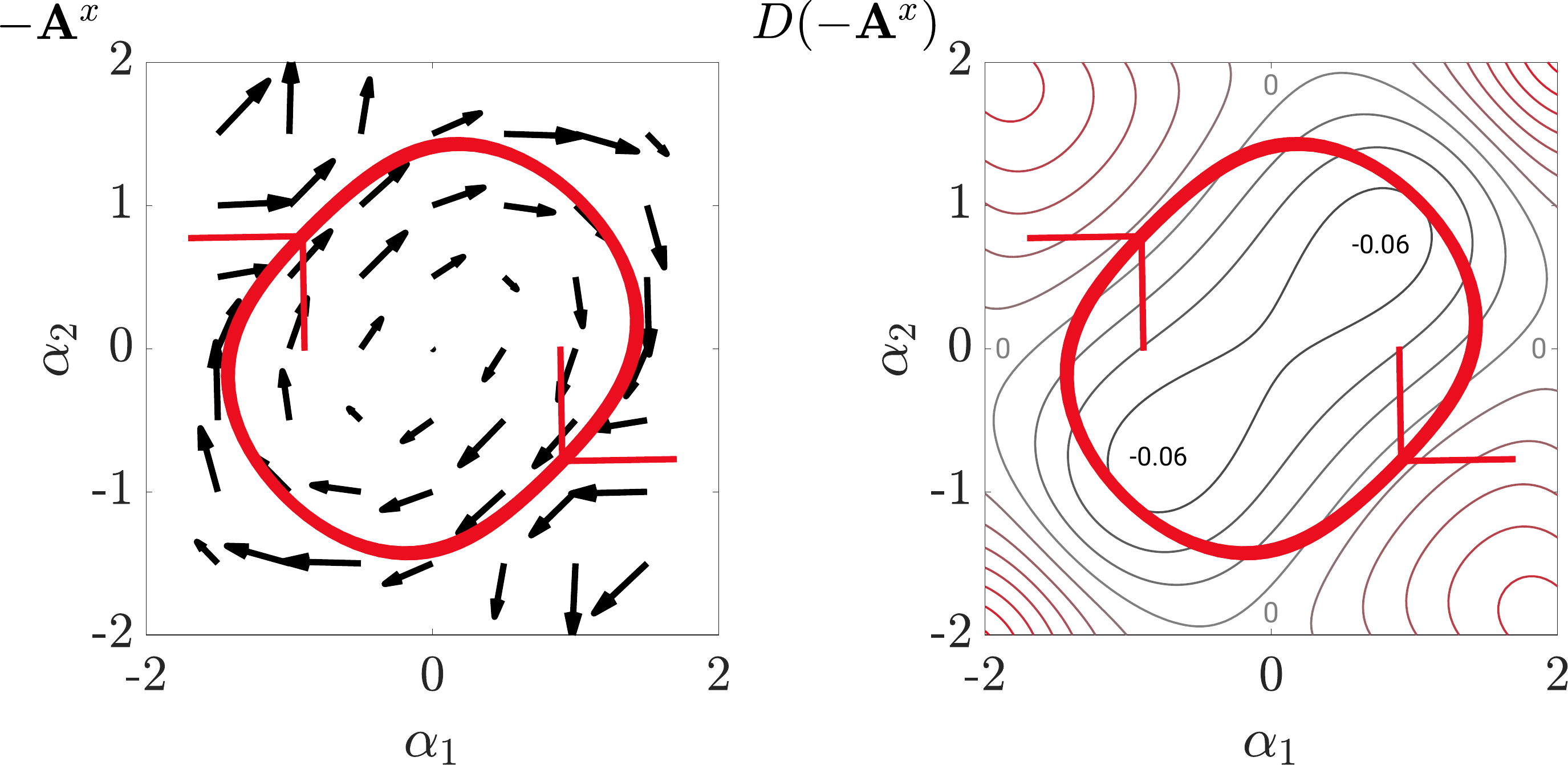}
\caption{The local connection (left) and CCF (right) of the forward ($x$) direction for a 3-link low Reynolds number swimmer in the optimal coordinates corresponding to the optimal gait solved by the proposed algorithm. The thickness of the gait corresponds to the pace, where a thicker pattern indicates slower movement.}
\label{fig:connccf}
\end{figure}

The solution to the system's equations of motion lies in the null space of the aggregated physics~\eqref{eq:nhc}, which can be rearranged into the \deleted{position velocity }reconstruction equation\footnote{\added{See Appendix~\ref{sec:overconstrained} for applying this method to overconstrained systems.}} \cite{hatton2011geometric}
\beq 
\fibercirc = \underbrace{-\inv{\left(\transpose{J}_{\fiber} \pfaffian J_{\fiber}\right)} \transpose{J}_{\fiber} \pfaffian J_{\base}}_{-\mixedconn}\basedot. \label{eq:kinrecon}
\eeq
The \emph{local connection} $\mixedconn$ can be visualized as a vector field in shape space as shown in the left panel of Fig.~\ref{fig:connccf}, which represents the system's position velocity resulting from the gait's shape velocity as it cycles through shape space.
\deleted{Note that this approach can also be applied to overconstrained systems, as discussed in Appendix~\ref{sec:overconstrained}.}

\subsection{Geometric Gait Analysis and Optimization}

The geometric mechanics community has developed methods that utilize the \emph{constraint curvature} of systems, which measures the net displacement resulting from periodic shape changes, to find the optimal gait \cite{murray1993nonholonomic, walsh1995reorienting, ostrowski1998geometric, melli2006motion, morgansen2007geometric, shammas2007geometric, avron2008geometric, hatton2013geometric, hatton2017kinematic, ramasamy2020geometry, hatton2022geometry}. The core principle is to consider that the net displacement $\gaitdisp$ over the gait cycle $\gait$ is the line integral of the exponential map of~\eqref{eq:kinrecon} along $\gait$. Therefore, the displacement caused by the gait only depends on the gait's path in the shape space.

More importantly, the resulting displacement can be approximated\footnote{This approximation is a generalized form of Stokes' theorem and a truncation of the Baker–Campbell–Hausdorf (BCH) series \cite{magnus1954exponential, hatton2015nonconservativity, bass2022characterizing}} as cBVI, which is the surface integral of the constraint curvature function (CCF) $D(-\mixedconn)$ of the local connection (its total Lie bracket) over the surface $\gait_{a}$ enclosed by the gait cycle $\gait$
\begin{eqalign}
\gaitdisp &= \ointctrclockwise_{\gait} -\fiber\mixedconn(\base) \span \span \\ 
&= \iint_{\gait_{a}} -\extd\mixedconn + \textstyle{\sum}\big{[}\mixedconn_{\fstidx},\mixedconn_{\sndidx>\fstidx}\big{]} + \text{higher-order terms} \span \span \\ 
&\approx \iint_{\gait_{a}} \underbrace{-\extd\mixedconn + \textstyle{\sum}\big{[}\mixedconn_{\fstidx},\mixedconn_{\sndidx>\fstidx}\big{]}}_{\text{$D(-\mixedconn)$ (total Lie bracket)}}, & (\text{cBVI}) \span \label{eq:ccf}
\end{eqalign}
\added{where $\extd\mixedconn$, the exterior derivative of the local connection (its generalized row-wise curl), measures the non-conservative component of net displacement, and the local Lie bracket $\sum\big{[}\mixedconn_{\fstidx},\mixedconn_{\sndidx>\fstidx}\big{]}$ measures the major non-commutative component}.\footnote{\added{Lie brackets use the net motion difference between different orders of translation and rotation to approximate non-commutativity, with its surface integral being the average non-commutativity over the gait cycle \cite{hatton2015nonconservativity, bass2022characterizing}.}}

$D(-\mixedconn)$ encodes the derivative of the displacement with respect to the gait geometry, providing the directions for gait optimization. For a system with a 2D shape space, the CCF can be visualized as a scalar function over the shape space, illustrating the effect of gait geometry on the resulting motions, as shown in the right panel of Fig.~\ref{fig:connccf}. In the case of a high-dimensional system with an $\sndqty$-dimentional shape space, the CCF becomes $\binom{\sndqty}{2}$ pairs of exterior products that encode the contributions of each pair of shape oscillations to the net displacement.

Based on cBVI and CCF, previous works in our lab have developed a geometric variational gait optimization algorithm \cite{ramasamy2016soapbubble}. This algorithm \added{finds} an equilibrium that maximizes the CCF flux captured by gait while minimizing the metric path length and pace. The optimization can be solved by extracting gradients from the CCF and gait geometry
\beq
\nabla_{\transpt} \fiber_{\gait} &\approx \ointctrclockwise_\phi\left(\sum^{\sndqty-1}_{\sndidx}\left(\nabla_{\transpt_{\perp_{\sndidx}}} \phi\right) D(-\mixedconn)_{\| \perp_{\sndidx}}\right),\label{eq:gaitopt}
\eeq
where $\|$ and $\perp_{\sndidx}$ represent the tangent direction and the $\sndidx$-th normal direction of the gait, respectively. \added{The algorithm can also be combined with various power consumption metrics to optimize gait efficiency} \cite{ramasamy2019geometry, hatton2022geometry, cabrera2023optimal}. Furthermore, the geometric gait optimization has been extended to systems with passive joints \cite{ramasamy2021optimal} and specified displacements \cite{choi2022optimal}.

However, the above geometric gait analysis and optimization still rely on the cBVI-based approximation in~\eqref{eq:ccf}. The accuracy of cBVI is significantly influenced by the proportion of higher-order terms that are neglected. When the system's locomotion has little non-commutativity, the higher-order terms become negligible \cite{hatton2015nonconservativity, bass2022characterizing}. This makes the cBVI an accurate predictor of the displacement of non-infinitesimal gaits and removes the dependence on the initial phase of the gait. In general, however, this condition of low non-commutativity exists only in gaits with small rotational amplitudes or in some special cases \cite{melli2006motion}. 

Previous works proposed coordinate optimization to identify a frame to optimally describe the overall motion of a multibody system and minimize the proportion of higher-order terms, ensuring an accurate cBVI-based approximation within a defined range \cite{hatton2011geometric, travers2013minimum}. Unfortunately, when applied to high-dimensional systems, coordinate optimization suffers from the curse of dimensionality. The next section presents a novel method to address this challenge and adapt coordinate optimization to high-dimensional systems.

\section{Gait-Based Coordinate Optimization}

In this section, we first review the factors that determine the proportion of higher-order terms ignored in geometric gait analysis. Subsequently, we revisit the definition of optimal coordinates in coordinate optimization. To address the challenge posed by the curse of dimensionality in current coordinate optimization methods, we propose a gait-based coordinate optimization method whose complexity grows linearly with the shape space dimension. Finally, we discuss the combination of the proposed gait-based coordinate optimization and geometric gait optimization.

\subsection{Characterization of Optimal Coordinates}

As described in \cite{ramasamy2020geometry, bass2022characterizing}, the cBVI is equivalent to a truncated Baker-Campbell-Hausdorff (BCH) series \cite{magnus1954exponential}. While the cBVI captures the first two terms of the series, the remaining higher-order terms still affect the net displacement. Fortunately, all remaining terms are \added{higher-order functions of the average and exterior derivative of local connections over the gait} \cite{ramasamy2020geometry, bass2022characterizing}. Therefore, by reducing the magnitude of the local connection, more information will be condensed into the first two terms.

The local connection of a new body frame, positioned relative to the original one by $\coordtrans(\base) \in \fiberspace$, is given by
\beq \label{eq:localconnectiontrans}
-\mixedconn_{\text{new}}=-\inv{\Adj_{\coordtrans}}\mixedconn + \inv{\coordtrans}\grad\coordtrans,
\eeq
where $\Adj_{\coordtrans}$ is an adjoint matrix mapping the original local connections to the new frame and $\inv{\coordtrans}\grad \coordtrans$ is the additional local connection resulting from the transformation. Optimal coordinates are defined by minimizing the total magnitude of local connections within the region of interest in shape space $\coordoptroi \subset \basespace$ \cite{hatton2010optimizing, hatton2011geometric}. The minimization object is computed as a multiple integral\deleted{ of the weighted Frobenius norm} over the region
\beq \label{eq:minpertubcoord}
\coordtrans^{*} = \argmin_{\coordtrans} \idotsint_{\base \in \coordoptroi}\left\|\mixedconn_{\text{new}}(\base)\right\|^{2}_{F, \coordoptweight} d\coordoptroi,
\eeq
where $\|\cdot\|_{F}$ is the Frobenius norm, and $\coordoptweight(r)$ is the weight assigned to each shape direction and each \added{grid}\deleted{mesh region within the region of interest}. 

Because $\mixedconn_{\text{new}}$ is a function of both $\coordtrans$ and $\grad\coordtrans$, the above minimization problem requires numerically solving a PDE. Previous works used a FEM-based approach to transform the PDE into a linear system and solve it \cite{hatton2011geometric, travers2013minimum}. However, this approach suffers from the curse of dimensionality because the complexity grows exponentially with the shape space dimension, which is nearly impossible for systems with high-dimensional shape spaces. Furthermore, because the FEM-based coordinate optimization needs to be solved for the entire shape space together, it requires information about the entire shape space, which makes it unsuitable for data-driven systems with only local information \cite{bittner2018geometrically}. \added{Although there have been attempts to solve this PDE in a mesh-free manner via random sampling in shape space \cite{chen2023integrable}, theoretically, achieving equivalent accuracy still requires complexity similar to that of FEM-based methods.}

\subsection{Gait-Based Coordinate Optimization}

\begin{figure}[!t]
\centering
\includegraphics[width=\linewidth]{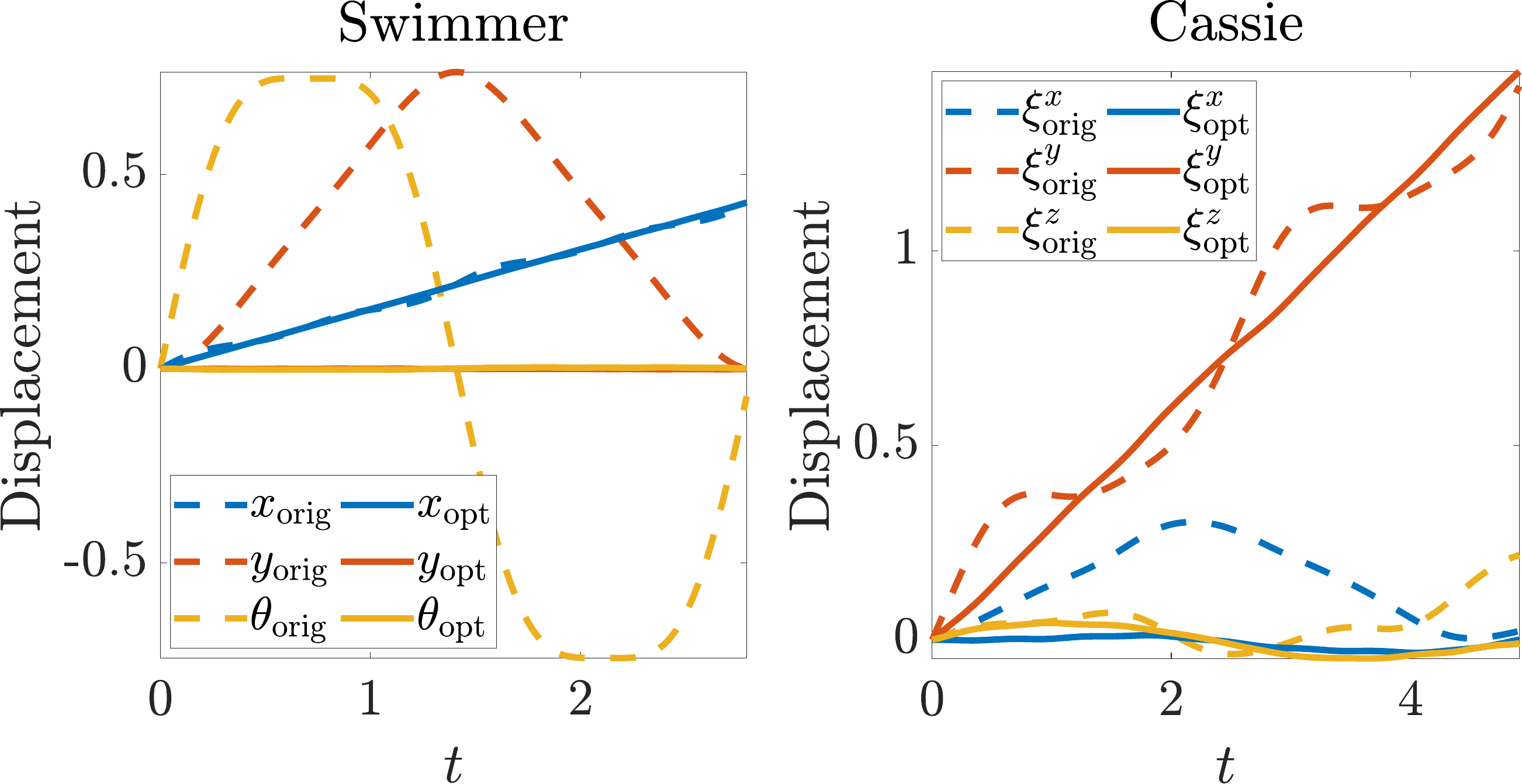}
\caption{The displacements resulting from the optimal gaits of the 12-link low Reynold number swimmer and the full model Cassie, measured respectively in their original frames attached to the central joint and torso, and in the optimal coordinates solved using the proposed method. Rigid body rotations are converted to exponential coordinates $\fiberexp$.}
\label{fig:mixed_displacement}
\end{figure}

In this paper, we exploit the insight from gait optimization~\eqref{eq:gaitopt} that the gradient of gait-induced net displacement depends only on the CCF within the gait neighborhood. This observation implies that we can perform coordinate optimization by focusing only on the gait neighborhood, and update coordinate optimization and gait optimization together. In this way, given a constant number of sampling points, we can achieve a complexity that grows linearly with the system dimensionality while maintaining good accuracy of the cBVI-based approximation. As shown in Fig.~\ref{fig:mixed_displacement}, the gait-based coordinate optimization successfully reveals the actual locomotion of the robot in the world and reduces oscillations and phase dependencies in displacement.

Specifically, we utilize the Kansa method \cite{kansa1990multiquadricsa, kansa1990multiquadrics, sarra2009multiquadric} to solve the coordinate optimization PDE problem~\eqref{eq:minpertubcoord} by sampling points along the gait $\gait$ and constructing an inverse quadratic radial basis function (RBF) at each sampled point to approximate each parameter of the transformation to the optimal coordinates
\beq
\coordtrans(\base) = \sum_{\base_{\trdidx}\in\gait}^{\samplingnum}\samplingweight_{\trdidx}\left(1 + \samplingconst^2\left\|\base - \base_{\trdidx}\right\|^{2}_{2}\right)^{-1},
\eeq
where $\base_{\trdidx}$ and $\samplingweight_{\trdidx}$ are the shape and the RBF weight of the $\trdidx$-th sampling point along the gait, $\samplingnum$ is the number of samples, and $\samplingconst$ is a hyperparameter.\footnote{In this paper, $\samplingnum$ is appropriately chosen to ensure accuracy and efficiency. $\samplingconst$ is selected to ensure that the resulting linear system is not ill-conditioned and the resulting transformation remains smooth.} The RBF is an infinitely smooth function and therefore a good candidate for an approximate numerical solution to the PDE problem. The minimization in~\eqref{eq:minpertubcoord} then becomes the gait-based coordinate optimization by sampling local connections along the gait $\gait$
\beq\label{eq:samplingminpertubcoord}
\samplingweight^{*} = \argmin_{\samplingweight} \sum_{\base_{\trdidx} \in \gait}^{\samplingnum}\left\|\mixedconn_{\text{new}}(\base_{\trdidx})\right\|^{2}_{F, \coordoptweight_{\trdidx}}.
\eeq
Because the variational gradient in~\eqref{eq:gaitopt} computes exterior product pairs between the tangent direction of the gait and all oscillation directions in shape space,\footnote{The exterior product of the tangent direction and itself is zero.} where the tangent direction appears $\sndqty+1$ times while the other directions appear only $1$ time, we match the weight $\coordoptweight_{\trdidx}$ to this scale.

For $\fiberspace = SE(2)$, we parameterize $\coordtrans = \rep\left(x, y, \theta\right)$, where $\rep$ is the parameterization function. On the other hand, for $\fiberspace = SO(3)$, because the rigid body rotation cannot be embedded in $\mathbb{R}^{3}$, we parameterize $\coordtrans = \exp\left(\coordtransexp\right)$, where $\coordtransexp\in\mathfrak{g} = so(3)$ is a Lie algebra element, and approximate~\eqref{eq:localconnectiontrans} by truncating the BCH series\footnote{\added{Note that when recovering the local connections at optimal coordinates from $\samplingweight^{*}$, we use~\eqref{eq:localconnectiontrans} for systems of $SO(3)$ group rather than the approximation~\eqref{eq:truncatedbch} \cite{bass2022geometric}.}} as in \cite{travers2013minimum}
\beq \label{eq:truncatedbch}
-\mixedconn_{\text{new}}\approx-\left(\matrixid-\coordtransexp\right)\mixedconn + \grad\coordtransexp,
\eeq
Given that the minimization problem in~\eqref{eq:samplingminpertubcoord} is a quadratic programming problem, one can easily apply the chain rule to compute the gradient of the minimization objective with respect to the RBF weights $\samplingweight$ as demonstrated in \cite{hatton2011geometric, travers2013minimum}. By setting the gradient to zero, we solve for the $\samplingweight^{*}$ that defines optimal coordinates
\beq \label{eq:coordgrad}
\samplingweight^{*} = \samplingweight\left|\nabla_{\samplingweight}\sum_{\base_{\trdidx} \in \gait}^{\samplingnum}\left\|\mixedconn_{\text{new}}(\base_{\trdidx})\right\|^{2}_{F, \coordoptweight_{\trdidx}} = 0\right..
\eeq
\deleted{Note that when recovering the local connections at optimal coordinates from $\samplingweight^{*}$, we use~\eqref{eq:localconnectiontrans} for systems of $SO(3)$ group rather than the approximation~\eqref{eq:truncatedbch}.}

\subsection{Combined Gait and Coordinate Optimization}

Considering the interdependence of the proposed gait-based coordinate optimization and geometric gait optimization, we perform both algorithms together. Specifically, in each iteration of gait optimization, we will first solve the coordinate optimization problem to update the optimal coordinates corresponding to the gait, so as to obtain accurate cBVI-based approximations and gradients. Because the algorithm we propose is a quadratic programming problem and can be easily solved by setting the gradient in~\eqref{eq:coordgrad} to zero, it imposes little additional computational burden on the gait optimization process.

We implemented the proposed algorithm in Matlab. The coordinate optimization problem was programmed to directly solve a system of linear equations based on~\eqref{eq:coordgrad}, while the geometric gait optimization was solved using the \textit{fmincon} function. For systems with 2D shape spaces, we took the circular gait covering the CCF peak region as an initial guess for gait optimization. For systems with high-dimensional shape spaces, we adopted the optimal gait of its reduced-order system as an initial guess.

\added{Note that the combined gait and coordinate optimization can be considered as the dual of whole-body motion planning \cite{dai2014wholebody} because both incorporate joint information of multi-body systems into motion planning. Additionally, the proposed coordinate optimization method can also be integrated with other control, planning, and estimation algorithms for broader applications, including bipedal robots walking on uneven terrain \cite{chen2023integrable, khripin2016natural}, to leverage optimal coordinates to enhance planning and feedback performance.}

\section{Simulation Results}

To test the proposed gait-based coordinate optimization method and its combination with geometric gait optimization, we evaluate them on two classes of high-dimensional systems, including the drag-dominated systems - low Reynolds number swimmers - modeled as 3, 6, 9, or 12 linkages, and the inertia-dominated system - Cassie in free fall - a bipedal robot with 10 DOFs.

\subsection{Low Reynolds Number Swimmer}

\begin{figure}[!t]
\centering
\includegraphics[width=\linewidth]{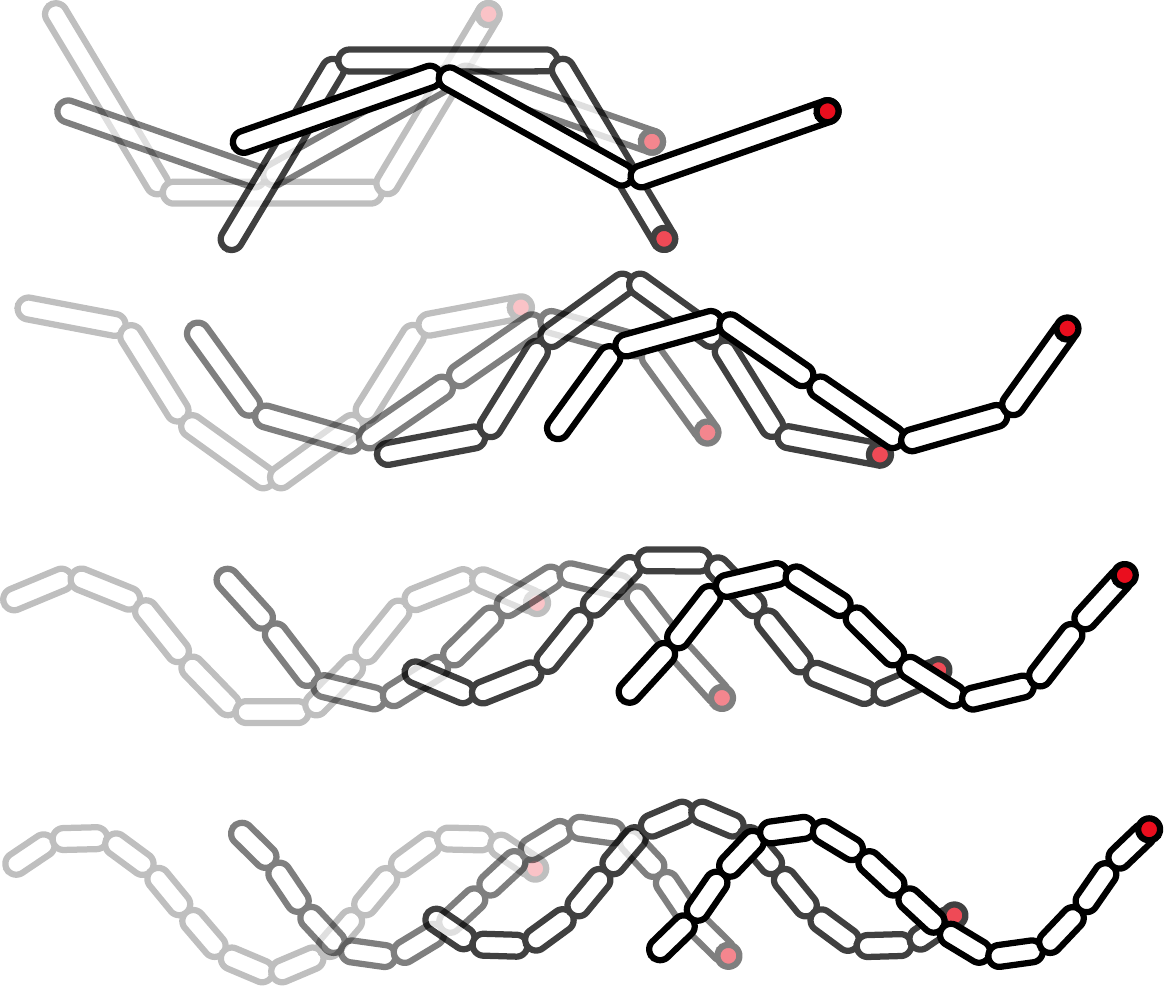}
\caption{Optimal gaits for each model of the low Reynolds number swimmer. The illustration shows the configuration of four quarter points within each gait's period. The displacement is scaled by the gait period to demonstrate efficiency. Each gait consumes unit power dissipation.}
\label{fig:swimmer_gait}
\end{figure}

\begin{figure}[!t]
\centering
\includegraphics[width=0.9\linewidth]{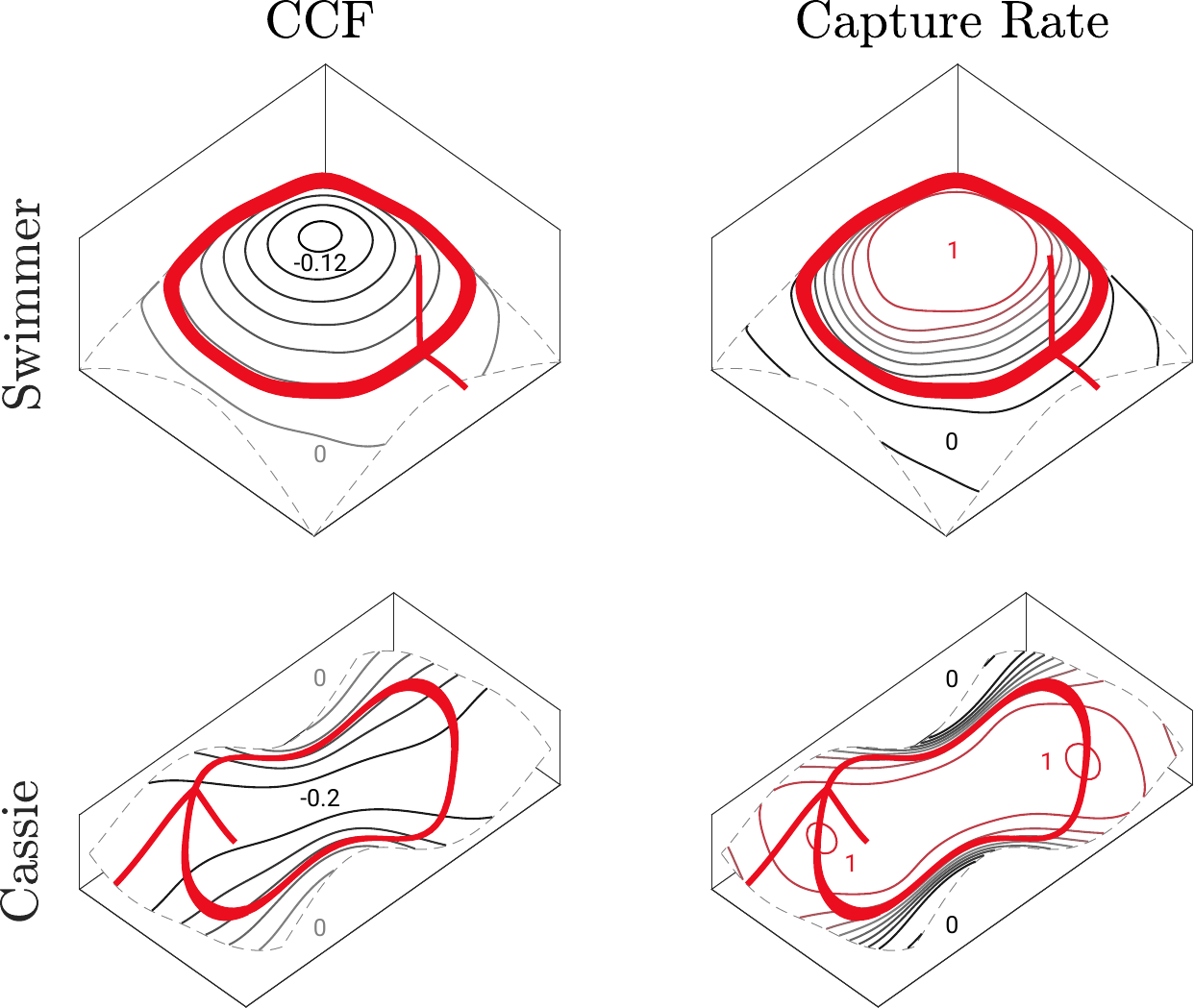}
\caption{Optimal gaits for the 12-link low Reynolds number swimmer (top) and the full model Cassie (bottom) in the 2D subspace with isometric 3D embedding. The contours show the CCF projected onto the subspace (left), and the CCF flux capture rate of the subspace relative to the maximum attainable CCF in the original shape space (right). Red, gray, and black represent positive, zero, and negative values of CCF, and represent 1, 0.5, and 0 for the capture rate. The thickness of the gait corresponds to the pace, where a thicker pattern indicates slower movement.}
\label{fig:mixed_optimality}
\end{figure}

The position of the low Reynolds number swimmer in the world is described by \added{the special Euclidean group, $SE(2)$}. Its dynamics are mainly determined by drag effects in~\eqref{eq:drag}, which implies that the total drag force and moment applied to the system will remain balanced \cite{hatton2013geometric}. The basic abstraction for a low Reynolds number swimmer is a 3-link system, but systems with more articulation can more accurately represent \added{continuum shapes and their} locomotion behavior in nature. We divide each link of the 3-link swimmers into 2, 3, and 4 parts, resulting in  6-, 9-, and 12-link systems, respectively. We apply the proposed combined gait-based coordinate optimization and geometric gait optimization to find the optimal gait that maximizes forward locomotion efficiency under the power dissipation metrics \cite{ramasamy2021optimal}.

Fig.~\ref{fig:swimmer_gait} shows the optimal gait for each model for low Reynolds number swimmers. Clearly, for the same power dissipation metric, systems with more articulation generate more efficient gaits, resulting in behaviors more closely resembling serpenoid motion observed in nature \cite{hirose1993biologically}. The left panel of Fig.~\ref{fig:mixed_displacement} shows the displacements of the 12-link swimmer measured in the original coordinates and the optimal coordinates determined by the proposed method. Compared to the original frames, the optimal coordinates reduce oscillations in other directions, especially rotation. These reduced oscillations minimize the non-commutativity and phase dependency of the system, thereby increasing the accuracy of cBVI and the effectiveness of gait optimization.

The top row of Fig.~\ref{fig:mixed_optimality} shows the optimal gait for the 12-link swimmer in a 2D subspace of the shape space. The subspace is computed by applying principal component analysis (PCA) to the optimal gait,\footnote{All 2D principal components in this paper can reconstruct over 95\% of the original gait.} and the three-dimensional embedding is constructed (via the Isomap algorithm~\cite{hatton2017kinematic}) such that path lengths through the surface are consistent with the metric-weighted cost of moving through the space. The contours in the left panel of Fig.~\ref{fig:mixed_optimality} show the CCF magnitude projected onto the isometric subspace. The contours in the right panel of Fig.~\ref{fig:mixed_optimality} show the ratio between the captured CCF flux per metric in the 2D subspace and the maximum attainable CCF flux per metric in the original space, providing a measure of alignment between the subspace and the optimal direction in the original space. Given the unit power dissipation metric, the optimal gait for the 12-link swimmer covers a CCF-rich region in the subspace. The subspace formed by the gait captures a large portion of the maximum attainable CCF flux. These characterizations together demonstrate the optimality of the gaits solved by the proposed algorithm.

\subsection{Cassie in Free Fall}

The position of the bipedal robot Cassie in free fall, including the aerial phase of running or jumping, can be described by the \added{special orthogonal group}, $SO(3)$, or its exponential coordinates $\xi$, within a limited range, considering that there are no external forces other than gravity. Its dynamics are determined by inertial effects, i.e.~\eqref{eq:momentum} with only 3D moments of inertia, 
meaning that momentum is conserved and remains zero throughout if starting from rest \cite{hatton2022geometry}. The closed-loop links on each leg are replaced with open-loop links of approximately equivalent inertia.\footnote{https://github.com/UMich-BipedLab/cassie\_description} We apply the proposed combined gait-based coordinate optimization and geometric gait optimization to find the optimal gait that maximizes the efficiency of angular locomotion in the pitch direction under the actuator force metrics \cite{hatton2022geometry}. We first consider a reduced-order model, where all joints are fixed in nominal positions except for the hip flexions and knees of each leg. These joints are controlled while maintaining a constant 180-degree phase shift between the two legs. We then use the optimal gait of the reduced-order model as an initial guess for the gait optimization of the full model of Cassie.

\begin{figure}[!t]
\centering
\includegraphics[width=\linewidth]{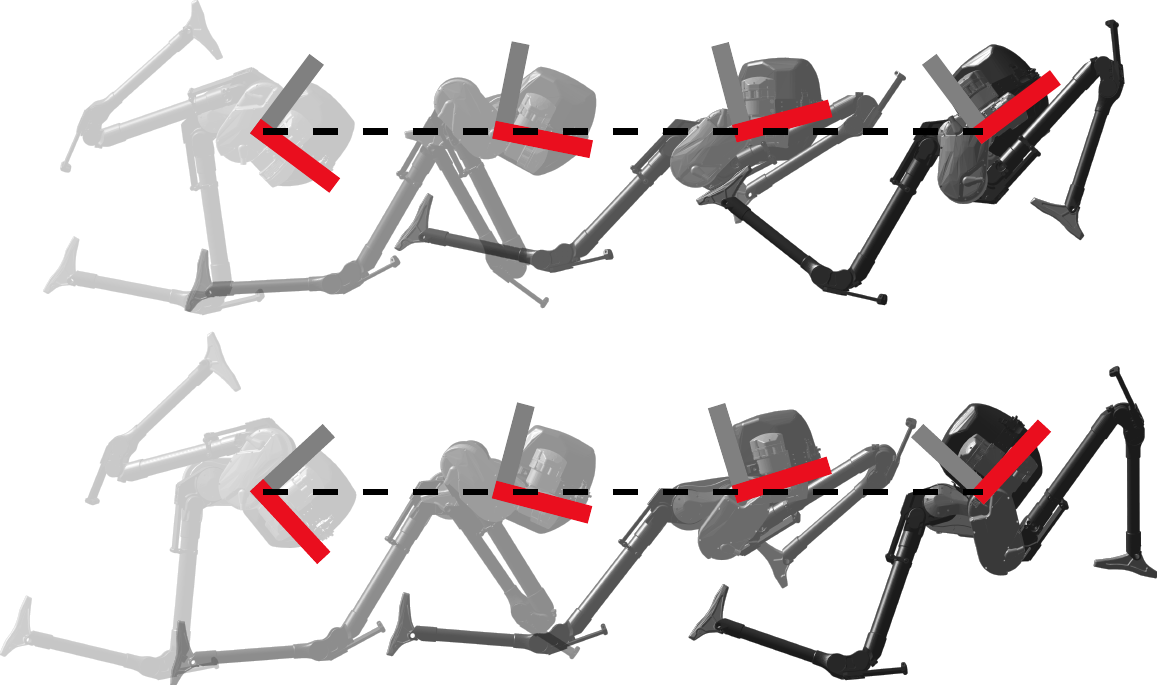}
\caption{Optimal gaits for the reduced-order model (top) and full model (bottom) of the free-falling Cassie. The illustration shows the configuration of four quarter points of a half period (swing and retraction of a single leg) of each gait. The rotation is scaled by the gait period to demonstrate gait efficiency. Each gait consumes a unit actuator force metric.}
\label{fig:cassie_gait}
\end{figure}

Fig.~\ref{fig:cassie_gait} shows the optimal gaits for both the reduced-order and the full model of free-falling Cassie. Similarly, for the same actuator force metric, \added{the full model produces more efficient gaits by utilizing additional DOFs to rotate the hip to adjust leg inertia during reciprocation.} The right panel of Fig.~\ref{fig:mixed_displacement} shows the displacements of the full model Cassie measured in the original coordinates and optimal coordinates determined by the proposed method. 
\added{Similar to the case of swimmers, the optimal coordinates greatly reduce oscillations in other rotational directions during the gait cycle, thereby improving the accuracy of the cBVI approximation and the effectiveness of gait optimization.}

The bottom row of Fig.~\ref{fig:mixed_optimality} shows the optimal gait for the full model Cassie in the isometric 3D embedding of the 2D subspace of shape space, similar to that of a swimmer. Note that the isometric embedding is based on the inertia matrix, while the actuator force metric considers the individual work of each actuator and therefore penalizes sharp turns more.
Given the unit actuator force metric, the optimal gait covers a CCF-rich region in the subspace. The subspace formed by the gait captures a large portion of the maximum attainable CCF flux, although the joint limits constrain further expansion of the gait to capture the peaks. These together demonstrate the optimality of the gaits solved by the proposed algorithm.

\section{Conclusion and Future Work}

This work proposes a gait-based coordinate optimization method that addresses the curse of dimensionality in the current geometric motion planning frameworks. We also propose a unified geometric representation of locomotion by generalizing various nonholonomic constraints into local metrics. By combining these two approaches, we take a step towards geometric motion planning for high-dimensional systems. We test our method in two classes of high-dimensional systems - low Reynolds number swimmers and free-falling Cassie - with up to 11-dimensional shape variables. Compared with the reduced-order model, the optimal gait obtained in the high-dimensional system shows better efficiency at the same power level. Furthermore, visualization in the isometric subspace provides a geometric optimality interpretation of the gaits. Future work includes 
incorporating higher-order terms in the BCH series into the gait analysis \cite{bass2022characterizing}, 
combining our approach with optimal gait families \cite{choi2022optimal} and data-driven dynamics \cite{bittner2018geometrically}, considering passive dynamics \cite{ramasamy2021optimal} and momentum drift \cite{shammas2007unified, yang2023geometric}, as well as integrating feedback control for hardware implementation \cite{mcisaac2003motion}.

\appendices

\section{Overconstrained Systems}
\label{sec:overconstrained}

For overconstrained systems with more nonholonomic constraints than the position space dimension, such as a kinematic snake with more than 3 links or a differential-drive vehicle with more than 2 wheels, the remaining constraints will impose constraints $\mathbf{G}$ on the shape velocity\footnote{$\ker(\cdot)$ and $\img(\cdot)$ are operators used to determine the kernel and image of a linear map. 
} \cite{dear2017locomotion, itani2021motion}
\beq
\underbrace{\transpose{\ker(\transpose{\jac_{\fiber}}\img(\pfaffian))}\transpose{\img(\pfaffian)}\pfaffian\begin{bmatrix}\jac_{\fiber} & \jac_{\base}\end{bmatrix}\transpose{\begin{bmatrix}-\transpose{\mixedconn} & \matrixid\end{bmatrix}}}_{\mathbf{G}}\basedot =0.
\eeq
Based on these constraints, we can determine the local connections in the permissible shape velocity direction by projecting them onto the feasible directions
\beq
\mixedconn_{\text{feasible}} = \mixedconn\ker(\mathbf{G})\transpose{\ker(\mathbf{G})}. \label{eq:proj}
\eeq
Our proposed local metric representation approach will yield the same feasible projections of local connections as alternative modeling methods such as choosing any arbitrary independent nonholonomic constraint of the same dimension as the position space, with the advantage of minimizing possible singularities in the local connections.

\bibliographystyle{IEEEtran}
\bibliography{ref}

\end{document}